\documentclass{article}

\usepackage{times}
\usepackage{graphicx} 
\usepackage{subfigure} 

\usepackage{natbib}

\usepackage{algorithm}
\usepackage{algorithmic}

\usepackage{hyperref}

\usepackage[accepted]{icml2017}

\icmltitlerunning{Inception Recurrent Convolutional Neural Network}

\begin{document} 

\twocolumn[
\icmltitle{Inception Recurrent Convolutional Neural Network for Object Recognition}
\icmlauthor{Md Zahangir Alom}{alomm1@udayton.edu}
\icmladdress{University of Dayton, Dayton, OH, USA}
\icmlauthor{Mahmudul Hasan}{mahmud.ucr@gmail.com}
\icmladdress{Comcast Labs, Washington, DC, USA}
\icmlauthor{Chris Yakopcic}{chris@udayton.edu}
\icmladdress{University of Dayton, Dayton, OH, USA}
\icmlauthor{Tarek M. Taha}{ttaha1@udayton.edu}
\icmladdress{University of Dayton, Dayton, OH, USA}

\icmlkeywords{deep learning, inception network, object recognition}
\vskip 0.3in
]

\begin{abstract} 
Deep convolutional neural networks (DCNNs) are an influential tool for solving various problems in the machine learning and computer vision fields. In this paper, we introduce a new deep learning model called an Inception-Recurrent Convolutional Neural Network (IRCNN), which utilizes the power of an inception network combined with recurrent layers in DCNN architecture. We have empirically evaluated the recognition performance of the proposed IRCNN model using different benchmark datasets such as MNIST, CIFAR-10, CIFAR-100, and SVHN. Experimental results show similar or higher recognition accuracy when compared to most of the popular DCNNs including the RCNN. Furthermore, we have investigated IRCNN performance against equivalent Inception Networks and Inception-Residual Networks using the CIFAR-100 dataset. We report about 3.5\%, 3.47\% and 2.54\% improvement in classification accuracy when compared to the RCNN, equivalent Inception Networks, and Inception-Residual Networks on the augmented CIFAR-100 dataset respectively. 
\end{abstract} 

\section{Introduction}
\label{Intro}
In recent years, deep learning using Convolutional Neural Networks (CNNs) has shown enormous success in the field of machine learning and computer vision. CNNs provide state-of-the-art accuracy in various image recognition tasks including object recognition \cite{1,2,3,4}, object detection \cite{5}, tracking \cite{6}, and image captioning \cite{7}. In addition, this technique has been applied massively in computer vision tasks such as video representation and classification of human activity \cite{8}. Machine translation and natural language processing are applied deep learning techniques that show great success in this domain \cite{9,10}. Furthermore, this technique has been used extensively in the field of speech recognition \cite{11}. Moreover, deep learning is not limited to signal, natural language, image, and video processing tasks, it has been applying successfully for game development \cite{12,13}. There is a lot of ongoing research for developing even better performance and improving the training process of DCNNs \cite{21,22,40,48,49}.

In some cases, machine intelligence shows better performance compared to human intelligence including calculation, chess, memory, and pattern matching. On the other hand, human intelligence still provides better performance in other fields such as object recognition, scene understanding, and more. Deep learning techniques (DCNNs in particular) perform very well in the domains of detection, classification, and scene understanding. There is a still a gap that must be closed before human level intelligence is reached when performing visual recognition tasks. Machine intelligence may open an opportunity to build a system that can process visual information the way that a human brain does. According to the study on the visual processing system within a human brain by James DiCarlo et al. \cite{14} the brain consists of several visual processing units starting with the visual cortex (V1), continuing through the extrastriate areas v2, v4, the PIT (Posterior Inferotemporal Area) cortex, and finally the AIT (Anterior Inferotemporal Area) cortex which is shown in Figure 1. It can be clearly seen that the visual cortex of the human brain processes information recurrently in different visual units. The recurrent connectivity of synapses in the human brain plays a big role in context modeling for visual recognition tasks \cite{14,19}.  
\begin{figure*}
	\centering
	\includegraphics[scale=0.76]{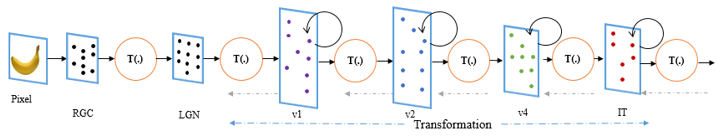}
	\caption{Visual information processing pipeline of the human brain, where v1 though v4 represent the visual cortex areas.}
	\label{fig:Fig_1}       
\end{figure*}
If we observe the architecture of recently developed DCNN models, most of them incorporate many components similar to that of the human visual information processing system for recognition tasks \cite{2,4}. 

However, the concept of recurrence in the visual cortex is only included in few DCNN models such as the Recurrent Convolutional Neural Network (RCNN) \cite{19}, and a CNN with LSTM for visual description \cite{34}. Additionally, Inception-V4 \cite{17}, and Residual \cite{18} architectures are popular among the computer vision community. The intension of most recently developed DCNNs is to use Inception and Residual networks to implement larger deep networks. As the model becomes larger and deeper, the computational parameters of the architecture are increased dramatically. As a result, training the model becomes increasingly complex and thus, more computationally expensive. It is very challenging to include a recurrent property within popular Inception architectures, but recurrence is essential for improving the overall training and testing accuracy with fewer computational parameters. Others are trying to implement bigger and deeper DCNN architectures like GoogleNet \cite{4}, or a residual network with 1001 layers \cite{16}   that achieves high recognition accuracy on different benchmark datasets. However, we are presenting an improved version of the DCNN model inspired by the information processing mechanisms of the human visual cortex, and recently developed some promising DCNN architectures like Inception-v4 \cite{17}, and RCNN \cite{19}.  Therefore, we call this model the Inception Recurrent Convolutional Neural Network (IRCNN). This model not only ensures better recognition accuracy with fewer computational parameters against the state-of-the-art DCNN architectures, but also helps to improve the overall training process of the deep learning approach. This proposed architecture generalizes both Inception networks and RCNN models. The contributions of this work are as follows: 
\begin{itemize}
	\item A new deep learning model called IRCNN is proposed.
	\item Experimental evaluation of the proposed learning model’s performance against different DCNN architectures on different benchmark datasets such as MNIST, CIFAR-10, CIFAR-100 and SVHN.
	\item Empirical investigation of the impact of the recurrent layer in the Inception Network.  
\end{itemize}
\section{Related work} 
\label{Related_work}
The deep learning revolution began in 1998 with \cite{20}. From then on, several different architectures have been proposed that have shown massive success using many different benchmark datasets including MNIST, SVHN, CIFAR-10, CIFAR-100, ImageNet, and many more.  Of the DCNN architectures, AlexNet \cite{2}, VGG \cite{3}, NiN \cite{21}, the All Convolutional Network \cite{22}, GoogleNet \cite{4}, Inception-v4 \cite{16,17}, and Residual Networks\cite{18} can be considered the most popular architectures due to their improved performance on different benchmarks for object classification.  In 2012, Alex Krizhevesky et al. proposed an improved version of a CNN model compared to LeNet \cite{20}, and won the ImageNet Large Scale Visual Recognition Challenge (ILSVRC) in 2012. This was a significant breakthrough in the field of machine learning and computer vision, as this was the first time a deep network outperformed the alternative approaches for visual recognition tasks. GoogleNet, or Inception-v1 \cite{4}, and the Residual Network \cite{18} won ILSVRC in 2014 and 2015 respectively. Inception architecture has become very popular in the deep learning and computer vision community, and has been refined in several ways. The improved versions of Inception networks with batch normalization \cite{48} (Inception-v2) were proposed by Ioffe et al. Later, an Inception network (Inception-v3) was proposed with factorization ideas in \cite{26}.
\begin{figure*}
	\centering
	\includegraphics[scale=0.86]{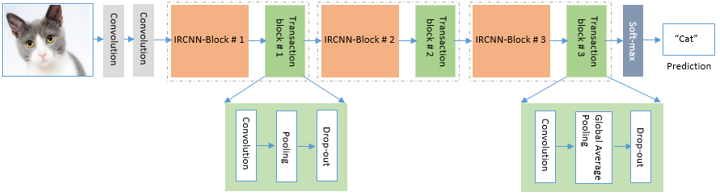}
	\caption{The overall operational flow diagram of the proposed Inception Recurrent Convolutional Neural Network (IRCNN), which consists of an IRCNN block, a transaction block, and a softmax layer.}
	\label{fig:Fig_2}       
\end{figure*}
In most cases, the improvement of deep learning approaches has been due to the development of the following components: Initialization techniques of DCNNs \cite{23}, new deep network architectures \cite{24,25}, optimized network structures (depending upon computational parameters) \cite{26,27}, deeper and wider deep networks \cite{28,29}, activation functions \cite{30}, and optimization methods for training DCNNs \cite{31,32}. Some researchers have been focusing on design alternatives that produce the same level of recognition accuracy as state-of-the-art architectures (like Inception-V4 with Residual Net \cite{17}) with fewer computational parameters \cite{27}.

Currently, most researches have been focused on improving the recognition accuracy of DCNN models. Very little research has been conducted on recurrent architectures within convolutional neural networks. The recurrent strategy is very important for context modeling from input samples. In 2015, Ming et al \cite{19} proposed a RCNN structure for the first time that dealt with object recognition tasks. The architecture consists of several blocks of recurrent convolutional layers followed by a max-pooling layer. In the second to last layer of the structure global max-pooling is used, which is followed by a soft-max layer at the end. This architecture showed state-of-the-art accuracy for object classification at the time \cite{19,33}. The Long-term Recurrent Convolutional Network (LRCN) was proposed for visual recognition and description by Jeff et al. This architecture uses a combination of two popular techniques, CNN and LSTM. The features are extracted through the CNN, and LSTM is applied to identify how features vary with respect to time. This model shows outstanding performance for visual description \cite{34}. From the above discussion, it can be concluded that DCNNs with improved architectures are showing enormous achievement when performing visual recognition tasks. 

\section{Inception-Recurrent Convolutional Neural Networks (IRCNN)}
\label{IRCNN}
The proposed architecture (IRCNN) builds on several recent developments in deep learning architectures, including Inception Nets \cite{26} and RCNNs \cite{19}. It tries to reduce the number of computational parameters, while providing better recognition accuracy. As shown in Figure 2, the IRCNN architecture consists of general convolution layers, IRCNN blocks, transaction blocks, and a softmax logistic regression layer. One of the most novel features of this work is the introduction of recurrence into the Inception module, as shown in the IRCNN block in Figure 3. The key feature of Inception-v4 [Szeged al et. 2015] is that it concatenates the outputs of multiple differently sized convolutional kernels in the inception block \cite{17,26}. Inception-v4 is a simplified version of Inception-v3 model, using lower rank filters for convolution. Inception-v4 however combines Residual concepts with Inception networks to improve the overall accuracy over Inception-v3. The outputs of the inception layers are added to the inputs of the Inception-Residual module. In this work, we utilize the inception concepts from Inception-v4.

The IRCNN block, performs recurrent convolution operations on different sized kernels (see Figure 3). In the recurrent structure, the inputs to the next time step are the sums of the convolutional outputs of the present time step and previous time steps. The same operations are repeated based on the time steps considered. As the input and output dimensions do not change, this is simply an accumulation of feature maps with respect to the time step considered. This helps to strengthen the extraction of the target features. As shown in Figure 3, an average pooling operation is applied before the recurrent convolution layer. In this particular pooling layer, $3\times3$ average pooling with stride $1\times1$ is applied by keeping the border size the same, resulting in output samples with the same dimensions as the inputs. The overlapping average pooling technique helps in the regularization of the network \cite{2}. 
\begin{figure}
	\centering
	\includegraphics[scale=0.65]{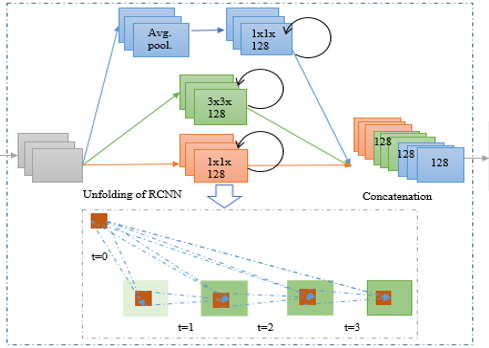}
	\caption{Inception-Recurrent Convolutional Neural Network (IRCNN) block with different convolutional layers containing different kernel sizes.}
	\label{fig:Fig_3}       
\end{figure}
The operations of each Recurrent Convolution Layer (RCL) in the IRCNN block are similar to operations in \cite{19}. To describe these operations, consider a pixel collated at $\left(i,j\right)$ of a particular input sample on the $k^{th}$ feature map in the RCL. This is the output $y_{ijk}(t)$ at time step $t$. The output can be expressed as:
\begin{equation}
y_{ijk}\left(t\right)={\left(w_k^f\right)}^T x_f^{\left(i,j\right)} \left(t\right)+{\left(w_k^r\right)}^T x_r^{\left(i,j\right)} \left(t-1\right)+b_k
\end{equation}
Here  $x_f^{\left(i,j\right)} \left(t\right)$ and $x_r^{\left(i,j\right)} \left(t-1\right)$ are the inputs for a standard convolutional layer and an RCL respectively.  $w_k^f$ and $w_k^r$ are the weights for the standard convolutional layer and the RCL respectively, and $b_k$ is the bias. The final output for the layer at time step $t$ is:
\begin{equation}
z_{ijk}\left(t\right)=f\left(y_{ijk}\left(t\right)\right)=max\left(0,y_{ijk}\left(t\right)\right)
\end{equation}
where $f$ is the standard Rectified Linear Unit (ReLU) activation function. The Local Response Normalization (LRN) function is applied to the outputs of the IRCNN-block\cite{2}.
\begin{equation}
y=norm\left(z_{ijk}\left(t\right)\right)
\end{equation}
The outputs of the IRCNN block with respect to the different kernel sizes and average pooling operations are defined as  $y_{1x1}\left(x\right)$,$y_{3x3}\left(x\right)$, and $y_{1x1}^p\left(x\right)$. The final output $y_{out}$ of the IRCNN-block can be described as:
\begin{equation}
y_{out}=y_{1x1}\left(x\right)\oplus y_{3x3}\left(x\right)\oplus y_{1x1}^p\left(x\right)
\end{equation}
Here $\oplus$ represents the concatenation operation with respect to the channel axis of the output samples. The outputs of the IRCNN-block become the inputs that are fed into the transaction layer.  

In the transaction block, three operations (convolution, pooling, and drop-out) are performed depending upon the placement of the block in the network. According to Figure 2, we have applied all of the operations in the very first transaction block; whereas in the second transaction block, we have only used convolution with dropout operations.  The third transaction block consists of convolution, global-average pooling, and drop-out layers. The global-average pooling layer is used as an alternative to a fully connected layer. There are several advantages of a global-average pooling layer. Firstly, it is very close in operation to convolution, hence enforcing correspondence between feature maps and categories. The feature maps can be easily interpreted as class confidence. Secondly, it does not need computational parameters, thus helping to avoid overfitting of the network. The softmax layer is used at the end of the IRCNN architecture. Late use of the pooling layer is advantageous because it increases the number of non-linear hidden layers in the network. Therefore, we have applied only two special pooling layers in this architecture. The max-pooling layers perform operations with a $3\times3$ patch and a $2\times2$ stride over the input samples. Since the non-overlapping max-pooling operation has a negative impact on model regularization, we used overlapped max-pooling for regularizing the network. This is very important for training a deep network architecture \cite{2}.  Special pooling is carried out with the max-pooling layer in the middle of the network (not all transaction blocks have pooling layer). Eventually, a global- average pooling layer is used at the very end before a softmax logistic regression layer.
To keep the number of computational parameters low compared to other traditional DCNN approaches like AlexNet \cite{2} and VGGNet\cite{3}, we have used only $1\times1$ and $3\times3$ convolutional filters in this implementation (inspired by the NiN \cite{21} and Squeeze Net \cite{27} models). There are significant benefits to using smaller sized kernels, such the ability to incorporate more non-linearity in the network. For example: we can use a stack of two $3\times3$ respective fields (without placing any pooling layer in between) as a replacement for one $5\times5$; and a stack of three $3\times3$ respective fields instead of a $7\times7$ \cite{3}. The benefit of adding a $1\times1$ filter is that it helps to increase the non-linearity of the decision function without having any impact on the convolution layer. Since the size of the input and output features do not change in the IRCNN blocks, it is just a linear projection on the same dimension with non-linearity added using a ReLU. We have used a dropout of 0.5 after each convolutional layer in the IRCNN-block. Finally, we have used a soft-max or a normalized exponential function \cite{35} layer at the end of the architecture. For an input sample $x$ and a weight vector $W$, and $K$ distinct linear functions the softmax operation can be defined for $i^{th}$ class as follows:
\begin{equation}
p(y=i|x) =\frac{e^{x^T} w_i}{\sum_{k=1}^{K} e^{x^T}w_k} 
\end{equation}
\section{Experiments}
We have evaluated the proposed IRCNN method (as well as several others for comparison) with a set of experiments on different benchmark datasets: MNIST \cite{36}, Cifar-10 \cite{37}, Cifar-100 \cite{37}, and SVHN \cite{38}. The entire experiment has been conducted on a Linux environment with Keras \cite{50} and Theano \cite{51} in the Backend running on a single GPU machine with an NVIDIA GTX-980.   
\subsection{Training Methodology}
In the first experiment, we trained the proposed IRCNN using the stochastic gradient descent (SGD) Nesterov technique with default initialization for deep networks found in Keras \cite{50}. We set the Nesterov momentum to 0.9 \cite{39}] and decay to $9.99\times e^-7$.  Second, we experimented with our proposed approach using the Layer-sequential unit-variance (LSUV) technique, which is a simple method for the initialization of weights in a deep neural network \cite{23}. We have also used a very recently proposed an improved version of the optimization function based on “Adam” that is called EVE \cite{31}. The following parameters are used for the EVE optimization function: the value of the learning rate $\left(\lambda\right)$ is $1\times e-4$, decay $\left(\gamma\right)$ is $1\times e-4$, $\beta_1=0.9$, $\beta_2=0.9$,  $\beta_3=0.9$, $k=0.1$, $K=10$, and $\epsilon=1\times e-08$. The $\left(\beta_1,\beta_2\right)\in[0,1)$ values are exponential decay rates for moment estimation in Adam. The 〖$\beta_3\in[0,1)$ is an exponential decay rate for computing relative changes. The  $k$, and $K$values are lower and upper thresholds for relative change and $\in$ is a fuzzy factor. It should be noted that we used the $l_2-norm$ with a value of 0.002 for weight regularization on each convolutional layer in the IRCNN block. 

In both experiments, we used ReLU activation functions. We have generalized the network with dropout (0.5). Only the horizontal flipping technique was applied when performing data augmentation. We trained the models for 350 epochs with a 128 batch size for CIFAR-10 and 100. During the training of MNIST and SVHN, we use 200 epochs with a mini-batch size of 128. For the impartial comparison, we have trained and tested against equivalent Inception networks and Inception residual networks (meaning our network contained the same number of layers and computational parameters). We describe these as the Equivalent Inception Network (EIN) and the Equivalent Inception Residual Network (EIRN).
\subsection{Results}
\subsubsection{MNIST}
MNIST is one of the most popular datasets for handwritten digits from 0-9 [36], the dataset contains $28\times28$ pixel grayscale images with 60,000 training examples and 10,000 testing examples. For this experiment, we trained the proposed model with two IRCNN convolution blocks (IRCNN-block 1 and IRCNN-block 2) and used the ReLU activation function. The model was trained with 60,000 samples and 10,000 samples were used for validation of the model. Eventually the trained network was tested with 10,000 testing examples. We obtained a test error of 0.32\% with the IRCNN and the SGD, and achieved about 0.29\% error for the IRCNN when initializing with LSUV \cite{23} and the EVE \cite{31} optimization function. This provided the best accuracy compared to the RCNN, as well as the other state-of-the-art networks. The summary of the classification accuracies is given in Table I.  No data augmentation techniques have been applied in this experiment on MNIST. On the contrary, global contract normalization and ZCA whitening were applied in the experiments using most of the mentioned models.  
\begin{figure}
	\centering
	\includegraphics[scale=0.70]{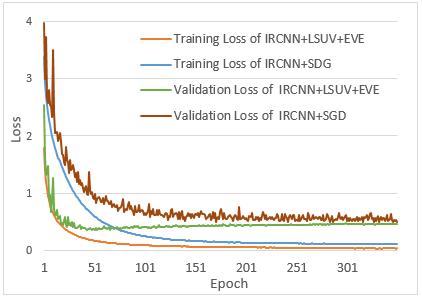}
	\caption{Training and validation loss of the IRCNN with SGD and LSUV+EVE on CIFAR-10.}
	\label{fig:Fig_4}       
\end{figure}
\begin{figure}
	\centering
	\includegraphics[scale=0.65]{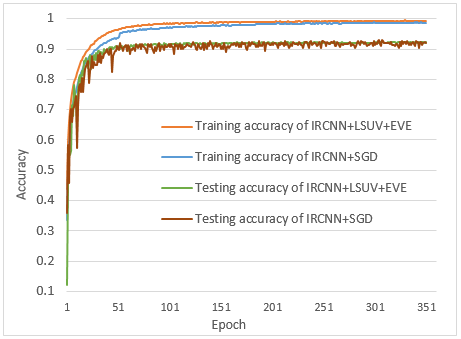}
	\caption{Training and validation accuracy of IRCNN with SGD and LSUV+EVE on CIFAR-10.}
	\label{fig:Fig_5}       
\end{figure}
\subsubsection{CIFAR-10}
CIFAR-10 is an object classification benchmark \cite{37} consisting of $32\times32$ color images representing 10 classes. It is split into 50,000 samples for training and 10,000 samples for testing. The experiment was conducted with and without data augmentation. The entire experiment was conducted on models similar to the one shown in Figure 2.  Using the proposed approach, we achieved about 8.41\% error without data augmentation and 7.37\% error with data augmentation using the SGD technique. These results are better than those of most of the recognized DCNN models stated in Table 1. 
Even better performance is observed from the IRCNN with LSUV \cite{23} as the initialization approach and EVE \cite{31} as the optimization technique. The results show around 8.17\% and 7.11\% error without and with data augmentation respectively. When comparing these results to those of the different models in Table 1, it can be observed that our proposed approach provides better accuracy compared to various advanced and hybrid models.The training and validation loss of the experiment on CIFAR-10 of this proposed model are shown in Figure 4.  Figure 5 shows training and validation accuracy of IRCNN with SGD and LSUV+EVE.
 
 \begin{table*}[t]
 	\caption{Testing errors (\%) on MNIST, CIFAR-10(C10), CIFAR-100(C100), and SVHN. Here $"+"$ indicates standard data augmentation using random horizontal flipping.IRCNN achieves lower testing errors in most of the cases indicate with bold.}
 	\label{sample-table2}
 	\vskip 0.15in
 	\begin{center}
 		\begin{small}
 			\begin{sc}
 				\begin{tabular}{lcccccr}
 					\hline
 					\abovespace
 					Methods&MNIST&C10&C10+&C100&C100+&SVHN\\
 					\hline
 					\abovespace
 					Maxout\cite{40}&0.45&11.6&9.38&-&38.57&2.47\\
 					NiN\cite{21}&0.47&10.4&8.81&35.68&-&2.35\\
 					DSN\cite{43}& 0.39&9.69&7.97&-&34.57&1.93 \\
 					CNN+Probout\cite{45}&-&9.39&-&-&38.14&2.39 \\
 					All-Conv.\cite{22}&-&9.08&7.25&-&33.71&-\\
 					Highway Net. \cite{44}&-& -&7.72&-&32.24&-\\
 					RCNN\cite{19}&0.31&8.69&\bf{7.09}&-&31.75&1.77\\
 					dasNet\cite{42}&-&-&9.22&-&33.78&-\\
 					FitNet\cite{46}&-&-&8.39&-&35.04&-\\
 					Drop-connect\cite{41}&1.12&-&9.41&-&-&1.94\\
 					CNN+Tree\cite{47}&-&-&-&-&36.85&-\\
 					\hline
 					\abovespace
 					\textbf{IRCNN+SGD}&0.32&8.41&7.37&34.13&31.22&1.89\\
 					\textbf{IRCNN+LSUV+EVE}&\bf{0.29}&\bf{8.17}& 7.11&\bf{30.87}&\bf{28.24}&\bf{1.74}\\
 					\hline
 				\end{tabular}
 			\end{sc}
 		\end{small}
 	\end{center}
 	\vskip -0.1in
 \end{table*}
\subsubsection{CIFAR-100}
This is another benchmarks for object classification from the same group (K and Hinton, 2009) \cite{37}. The dataset contains 60,000 (50,000 for training and for 10,000 testing) color $32\times32$ images, and it has 100 classes. We have used SGD and LSUV \cite{23} as the initialization approach with the EVE optimization technique \cite{31} in this experiment. The experimental results are shown in Table 1. In both cases, the proposed technique shows state-of-the-art accuracy compared with different DCNN models. 
\begin{figure}
	\centering
	\includegraphics[scale=0.72]{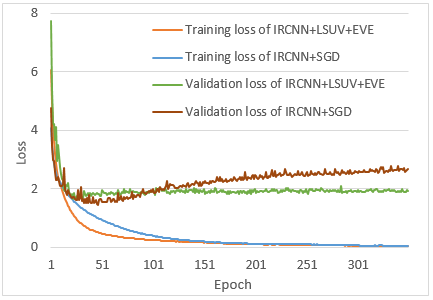}
	\caption{Training and validation loss of IRCNN with SGD and LSUV+EVE on CIFAR-100.}
	\label{fig:Fig_6}       
\end{figure}
IRCNN+SGD shows about 34.13\% testing errors without data augmentation and 31.22\% classification errors with data augmentation. In addition, this models achieved around 30.87\% and only 28.24\% errors with SGD and LSUV+EVE on augmented dataset. This is the highest accuracy achieved in any of the deep learning models summarized in Table 1. For augmented datasets, we have achieved 71.76 \% recognition accuracy with LSUV+EVE, which is about a 3.51\% improvement compared to RCNN\cite{19}. 
Figure 6 shows the training and validation loss of IRCNN for both experiments on CIFAR-100 with data augmentation(with initialization and optimization). It is clearly shown that the proposed model has lower error in the both experiments, showing the effectiveness of the proposed IRCNN learning model. The training and testing accuracy of the IRCNN with LSUV and EVE are shown in Figure 7.
\begin{figure}
	\centering
	\includegraphics[scale=0.75]{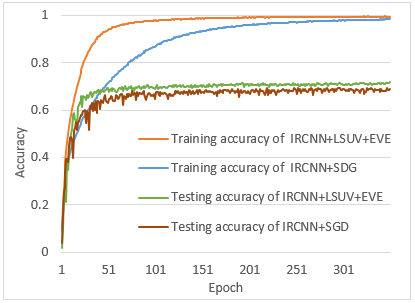}
	\caption{Training and validation accuracy of IRCNN with SGD and LSUV+EVE on CIFAR-100.}
	\label{fig:Fig_7}       
\end{figure}
\subsubsection{Street View House Numbers (SVHN)}
SVHN (Netzer et al. 2011) is one of the most challenging datasets for street view house number recognition \cite{38}. This dataset contains color images representing house numbers from Google Street View. In this experiment, we have considered the second version, which consists with $32\times32$ color examples. There are 73,257 samples are in the training set and 26,032 samples in testing set. In addition, this dataset has 531,131 extra samples that are used for training purposes. As single input samples of this dataset contain multiple digits, the main goal is to classify the central digit. Due to the huge variation of color and brightness, this dataset is much for difficult to classify compared to the MNIST dataset. In this case, we have experimented with the same model as is used in CIFAR-10 and CIFAR-100.  We have used the same preprocessing steps applied in the experiments of RCNN \cite{19}. The experimental results show better recognition accuracy, as shown in Table 1. We have obtained around 1.89\% testing errors with IRCNN+SGD and 1.74\% errors with IRCNN+LSUV+EVE respectively.  It is noted that Local Contract Normalization (LCN) is applied during experiments of MaxOut \cite{40}, NiN \cite{21}, DSN \cite{43}, and Drop Connect \cite{41}. The drop connection results in \cite{41} are based on the average performance of five networks.

\subsection{Impact of recurrent layers }
The proposed architecture also performs well when compared to traditional architectures. One LSUV with a traditional DCNN architecture is called FitNet4, and it only achieved 70.04\% classification accuracy with data augmentation using mirroring and random shifts for CIFAR-100 \cite{23}. On the other hand, we have only applied random horizontal flipping for data augmentation in this implementation and achieved about 1.72\% better recognition accuracy against FitNet4 \cite{23}.
\begin{figure}
	\centering
	\includegraphics[scale=0.66]{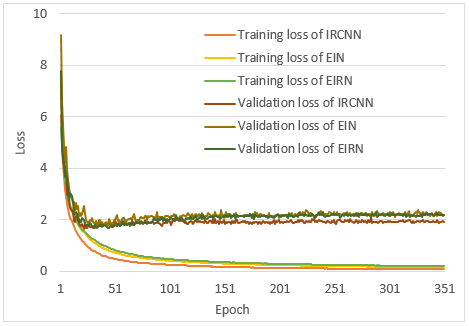}
	\caption{Training and validation loss of IRCNN, EIN, and EIRN on CIFAR-100.}
	\label{fig:Fig_8}       
\end{figure}
For an impartial comparison with the EIN and EIRN models, we have implemented the Inception network with the same number of layers and parameters as in the transaction and Inception-block. Instead of using recurrent connectivity in the convolutional layers, we used sequential convolutional layers for the same time-step with the same kernels. During the implementation of EIRN, we only added residual connection in the Inception-Residual block, where the inputs of the Inception-Residual block are accumulated with the outputs of that particular block.
\begin{figure}
	\centering
	\includegraphics[scale=0.70]{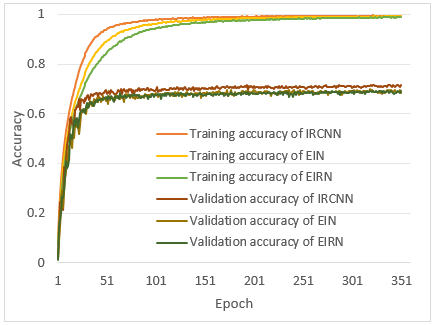}
	\caption{Training and validation accuracy for IRCNN, EIN, EIRN on CIFAR-100.}
	\label{fig:Fig_9}       
\end{figure}
In this case, all of the experiment have been conducted on the augmented CIFAR-100 dataset \cite{37}. The model loss and accuracy for both training and validation phases are shown in Figures 8 and 9 respectively. From both figures, it can be clearly observed that this proposed model shows lower loss and the highest recognition accuracy compared with EIN and EIRN, proving the effectiveness of the proposed model. It also demonstrates the advantage of recurrent layers in Inception networks. The testing accuracy of IRCNN, EIN, and EIRN are shown in Figure 10. It can be summarized that our proposed model of IRCNN shows around 3.47\% and 2.54\% better testing accuracy compared to EIN and EIRN respectively. 
\begin{figure}
	\centering
	\includegraphics[scale=0.60]{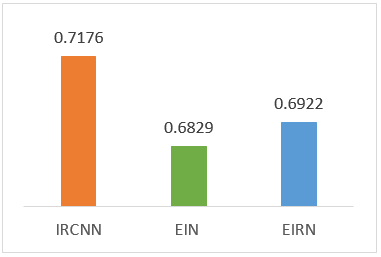}
	\caption{Testing accuracy of proposed IRCNN model against EIN and EIRN on augmented dataset of CIFAR-100.}
	\label{fig:Fig_10}       
\end{figure}
\subsection{Evaluation}
From the above empirical evaluations, it can be concluded that this proposed technique provides better recognition accuracy compared to different deep learning models in most of the cases, demonstrating the precision of the proposed deep learning model. This model also shows better recognition performance with the same number of computational parameters $(\sim 3.12M)$ against the EIN and EIRN models. Furthermore, if we observed the figures for model loss and accuracy, it can be clearly seen that this proposed model demonstrates less loss with better recognition accuracy. We have also empirically evaluated the rate of convergence of our proposed IRCNN algorithm compared with traditional EIN and EIRN models. The proposed model converged earlier with much lower model loss compared to EIN and EIRN. The computational cost (in seconds) per epoch of this IRCNN,EIN, and EIRN models for different benchmark datasets are shown in Table 2.
\begin{table}[t]
	\caption{Computation time per Epoch for IRCNN, EIN and EIRN models }
	\label{sample-table1}
	\vskip 0.15in
	\begin{center}
		\begin{small}
			\begin{sc}
				\begin{tabular}{lcccr}
					\hline
					\abovespace
					Model & Dataset & Time (in Sec.)  \\
					\hline
					\abovespace
					IRCNN    & MNIST & 112 \\
					IRCNN & CIFAR-10    & 418 \\
				    IRCNN & CIFAR-100    & 422 \\
					IRCNN & SVHN    & 610 \\
					EIN & CIFAR-100    & 425 \\
					EIRN & CIFAR-100    & 426 \\
					\hline
				\end{tabular}
			\end{sc}
		\end{small}
	\end{center}
	\vskip -0.1in
\end{table}
\subsection{Introspection}
In this implementation, we have augmented data applying only random horizontal flipping techniques whereas other models published results with more data augmentation with transaction, central crop, and ZCA. This proposed model will provide better recognition accuracy when using datasets with additional augmentation.Due to hardware constrains, we were not able to experiment on massive scale implementations of IRCNN. This architecture will probably provide even better classification accuracy with large networks on the same datasets. The large scale implementation of proposed IRCNN model with advanced components such as LSUV, EVE, and Exponential Linear Unit (ELU) \cite{30} will likely provide further improved recognition on CIFAR-10 and CIFAR-100 datasets. 
\section{Conclusion and Future works} 
In this paper, we have proposed a new architecture: Inception Recurrent Convolutional Neural Network (IRCNN) for object recognition where we have utilized the power of recurrent techniques for context modulation with the architecture of Inception networks. The experimental results show the promising recognition accuracy compared with different state-of-the-art Deep Convolutional Neural Networks (DCNN) models on different benchmark datasets such as MNIST, CIFAR-10, CIFAR-100, and SVHN. However, when the proposed IRCNN architecture is initialized with LSUV initialization technique, and optimization function of EVE, it achieved an object recognition accuracy of 71.76\% on the CIFAR-100 dataset. This is about a 3.5\% improvement with respect to RCNN \cite{19}. In addition, this architecture accelerates the training procedure, which is a concerning issue right now for training large scale deep learning approaches. Furthermore, we empirically investigated our model and determined that it outperforms against both the equivalent model of the Inception Networks and the Inception-Residual Networks. 
In the future, we would like to improve this model and experiment with large scale implementation using the teacher-student paradigm (Net2Net) on the ImageNet dataset \cite{52}. In addition, the propose IRCNN will be tested with advanced activation functions such as ELU \cite{30}.Furthermore, from our observation, this new architecture would be able to model context in input videos, which is another future direction for this work.

 

\bibliography{ICML_references}
\bibliographystyle{icml2017}

\end{document}